# Performing lexical quality estimation of a large OCRed historical Finnish newspaper collection with scarce resources

Kimmo Kettunen
National Library of Finland, DH Projects

**Abstract**

The National Library of Finland has digitized and made available historical newspapers published in Finland since the late 1990s. The present size of the open journalistic collection is about 13 million pages, mainly in Finnish and Swedish, the two official languages of Finland. The open part of the collection covers years 1771–1929. The National Library's Digital Collections are offered via the *digi.kansalliskirjasto.fi* web service, also known as *Digi.*

In this study we analyze word level quality of the Finnish collection between 1771 and 1910 (Bremer-Laamanen 2014; Kettunen et al. 2014). The Finnish part of the collection consists of ca. 938 275 pages and about 2.40 billion words.

Quality of OCRed collections is an important topic in digital humanities, as it affects general usability and searchability of collections. There is no single available method to assess quality of large collections, but different methods can be used to approximate quality. This paper discusses different corpus analysis style methods to approximate overall lexical quality of the Finnish part of the Digi collection. Methods include usage of parallel samples and word error rates, usage of morphological analysers, frequency analysis of words and comparisons to comparable edited lexical data. Our aim in the quality analysis is twofold: firstly to analyse the present state of the lexical data and secondly, to establish a set of assessment methods that build up a compact procedure for quality assessment after e.g. re-OCRing or post-correction of the material. In the discussion part of the paper we shall synthesise results of our different analyses.

## 1. Introduction

Newspapers of the 19th and early 20th century were mostly printed in the Gothic (Fraktur, blackletter) typeface in Europe. It is well known that the typeface is difficult to recognize for OCR software (Holley 2008; Furrer and Volk 2011; Volk et al. 2011). Other aspects that affect the quality of the OCR recognition are the following, among others (cf. Holley 2008; Klijn 2008, for a more detailed list):

- quality of the original source and microfilm
- scanning resolution and file format
- layout of the page
- OCR engine training
- unknown fonts
- etc.

As a result of these difficulties scanned and OCRed document collections have a varying amount of errors in their content. A quite typical figure is that of the 19th Century Newspaper Project of the British Library (Tanner et al. 2009): they report that 78 % of the words in the collection are correct. This kind of quality is not very good, but quite realistic. The amount of errors depends heavily on the period and printing form of the original data. Older newspapers and journals are more difficult for OCR; newspapers from the early 20th century are easier (cf. for example data of Niklas 2010 that consists of a 200 year period of The Times of London from 1785 to 1985). There is no clear measure of the amount of errors that makes OCRed material useful or less useful for some purpose, and the use purposes and research tasks of the users of digitized material vary hugely (Traub et al. 2015). A linguist who is interested in the forms of words needs as errorless data as possible; a historian who interprets texts on a more general level may be satisfied with text data that has more errors. In any case, the quality of the OCRed word data is of crucial importance.

Digital collections may be small, medium sized or large and different methods of quality assessment are useful or practical for different sizes of collections. Smallish and perhaps even medium sized collections may be assessed and corrected intellectually, by human inspection (cf. Strange et al. 2014). When size of the collection increases, human inspection becomes impossible, or it can only be used to assess samples of the collection. In our case, the size of the collection makes comprehensive human inspection impossible: over 938 000 newspaper *pages* of varying quality cannot be assessed by human labour.

Thus quality assessment of OCRed collections is most of the times *sample-based*, as in the case of the British Library's 19th Century Newspapers Database (Tanner et. al. 2009)[1]. A representative part of the collection is assessed e.g. by using a parallel digital clean collection, when such is available or can be produced cost effectively. Word and character level comparisons can then be made and error rates of the OCRed collections can be reported and compared. Holley (2008) gives the following, mainly practical, OCR quality accuracy figures and quality estimations that are based on discussions with OCR contractors and academic librarians:

| | | | |
|---|---|---|---|
| **Good** | OCR accuracy | 98–99% accurate | (1–2% of OCR incorrect) |
| **Average** | OCR accuracy | 90–98% accurate | (2–10% of OCR incorrect) |
| **Poor** | OCR accuracy | below 90% accurate | (more than 10% of OCR incorrect)[2] |

Another, fully automatic possibility to assess quality of the collection is usage of digital dictionaries Niklas (2010), for example, uses dictionary look-up to check the overall word level quality of The Times of London collection from 1785 to 1985 in his OCR post-correction work. This kind of approach gives a word accuracy approximation for the data (Strange et al. 2014).

Unfortunately usage of dictionaries suits only languages like English that have only little inflection in words and thus the words in texts can be found in dictionaries as dictionary entries. A heavily inflected, morphologically complex language like Finnish needs other means: full morphological analysis of the material is needed for this type of language. We shall discuss this approach with our material later on.

Some other methods could also be used. Baeza-Yates and Rello (2012) suggest a simple spelling error based look-up method to evaluate lexical quality of web content, based on the original idea of Gelman and Barletta (2008). We believe that this method might also be useful in analysis of our data, but we are not able to discuss its possible use at present.

[1] "To discover the actual OCR accuracy of the newspaper digitization program at the BL we sampled a significant proportion (roughly 1%) of the total 2 million plus pages..." This kind of approach where a clean parallel data for the OCRed sample is produced in house or by a contractor, is beyond our means.

[2] Unfortunately it is not clear, whether accuracy here is on character or word level, but for the sake of discussion we'll suggest that the figures are word level accuracy figures, as even high character level accuracy can mean quite low word level accuracies (Tanner et al. 2009).

Statistical methods are used many times in corpus linguistics (e.g. Kilgariff 2001), especially when different corpora are compared. In our case usage of statistics is perhaps not very feasible at present stage, where we need to establish a base level quality approximation for our data. Very high level statistical methods, where the status of the OCRed historical documents are profiled (e.g. Reffle and Ringlstetter 2013) show also promise, but their use at the present is beyond our means.

This paper discusses different corpus analysis style methods to approximate overall lexical quality of the Finnish part of the Digi collection 1771-1910. Methods include usage of parallel samples and word error rates, usage of automatic morphological analysers, frequency analysis of words and comparisons to comparable edited lexical data. Our aim in the quality analysis is twofold: firstly to analyse the present state of the lexical data and secondly, to establish a set of assessment methods that build up a compact procedure for quality assessment after e.g. re-OCRing or post-correction of the material.

## 2. Quality assessment of the Digi

### 2.1 Data and tools

The journalistic collection of the National Library of Finland has been digitized and OCR'ed since the year 1998. The OCR engine used for optical character recognition has been ABBYY FineReader, starting from version 6 onwards. The presentation system for on-line access to the collection is docWorks. The National Library of Finland has digitized and made available the historical newspapers published in Finland between 1771 and 1910 (Bremer-Laamanen 2014; Kettunen et al. 2014). This collection contains approximately 1.95 million pages in Finnish and Swedish. Finnish part of the collection consists of about 2.40 billion words and over 938 000 pages. The National Library's Digital Collections are offered via the digi.kansalliskirjasto.fi web service, also known as Digi. An open data package of the whole collection was released in early 2017 (Pääkkönen et al., 2016).

We started assessment of the quality of the Digi in 2014. Part of this work has been described in Kettunen et al. (2014) and Kettunen (2015). These publications describe mainly first post-correction trials of the Finnish newspaper material. To that effort we set up semi-automatically seven smallish parallel corpora (ca. 212 000 words) upon which post-correction trials were performed. Results of the evaluation showed that the quality of the evaluation data varied from about 60 % word accuracy at worst to about 90 % accuracy at best, the mean being about 75 % word accuracy. The evaluation samples, however, were small, and on the basis of the parallel corpora it is hard to estimate what the overall quality of the data is. Scarce availability of edited 19th century parallel newspaper material (Lauerma, 2012) makes this approach also hard to continue any further and there are no resources to set up larger parallel data for evaluation purposes by ourselves. Thus another type of approach is needed.

Since the first trials we have further worked on lexical level with our data. In winter 2015 we extracted the database of the 1771–1910 Digi collection and extracted the words from the page texts of the dump. Punctuation of the text was discarded in the dump but distinction between lower and upper case letters was kept in the resulting word lists[3].

We got two different word lists: the first and smaller one consists of all the Finnish newspaper and journal word material up to year 1850. It has about 22 million word form tokens, which is less than 1 % of the whole data. The second and more interesting word list consists of the Finnish words in the newspapers during the period 1851–1910 and it contains about 2.39 billion word form tokens. As the main volume of the lexical data of the collection is in the 1851–1910 section of the corpus, we shall concentrate mainly on the analysis of this part of the corpus, but will show also some results of the time period of 1771–1850.

As far as we know there is no single method or ICT system available that could be used for analyzing the quality of word data in a very large historical newspaper collection.[4] Thus we ended up in using a few simple ways to approximate quality of our data. Firstly we analyzed all the words of the index with two modern Finnish morphological analyzers, commercial FINTWOL[5] and open source Omorfi[6]. As there is no fully developed morphological analyzer of historical Finnish available, this is the only possible way to do morphological analysis for the data[7]. A typical morphological analyzer consists of a rule component and a large lexicon (Pirinen, 2015). If the analyzer can relate an input word after application of rules to a base form or forms in its lexicon, it has successfully recognized/analyzed the word. We ran our data through the analyzers and counted how many of the words were recognized or unrecognized by the analyzers. Naturally the number of unrecognized words contains both historically/dialectically unknown words for the modern Finnish analyzers (out-of-vocabulary, OOV, includes also words in foreign language) **and** OCR errors. A positive recognition does not also guarantee that the word was what it was in the original text. However, when the figures of our analyzed data are compared to analyses of existing edited dictionary and other data of the same period, we can approximate, what amount of our data could be OCR errors.

[3] This means that notion *word* has been defined in the indexing phase of the collection as a character string separated by spaces on left and right side.

[4] EU Project IMPACT, http://www.digitisation.eu/, has produced lots of tools for different purposes, but there does not seem to be a suitable tool for this purpose. Nor does CLARIN's inventory (http://www.clarin.eu/content/language-resource-inventory) have any suitable software. System described in Reffle and Ringlstetter (2013) could be suitable, but it would need modifications to handle complex morphology of Finnish.

[5] http://www2.lingsoft.fi/doc/fintwol/; We use FINTWOL's version 1999/12/20.

[6] https://github.com/flammie/omorfi; We use omorfi-analyse version 0.1, dated 2012 and 0.2 dated Oct 14, 2014

[7] A statistical lemmatizer described in Loponen and Järvelin (2010) might also be suitable for our purposes, but unfortunately this software is not available. Morfessor type (http://www.cis.hut.fi/projects/morpho/) unsupervised morph segmenting software, on the other hand, is useless for our purposes.

Secondly, we made frequency calculations of the word data and took different samples out of that data for further analysis with the morphological analyzers. These analyses give a more detailed picture of the data.

Table 1 shows initial recognition rates of all the word tokens and types in the Digi with the two morphological analyzers. In Table 2 we results with a more history aware version of Omorfi[8] (we call this HisOmorfi) and a later version of Omorfi, v. 0.2, are given.

| **Collection** | **Number of words** | **Recognized by Omorfi 0.1** | **Recognized by FINTWOL** |
|---|---|---|---|
| Digi up to 1850 tokens | 22.8 M | 65.6 % | 65.2 % |
| Digi 1851-1910 tokens | 2.385 G (2 385 349 514) | 69.3 % (1 652 668 099) | N/A |
| Digi up to 1850 types | 3.24 M | 15.6 % | 14.9 % |
| Digi 1851-1910 types | 177.3 M | 3.8 % | 3.5 % |

**Table 1.** Recognition rates for word types and tokens of Digi: FINTWOL and OMorfi 0.1

| **Collection** | **Number of words** | **Recognized by Omorfi 0.2** | **Recognized by HisOmorfi** |
|---|---|---|---|
| Digi up to 1850 tokens | 22.8 M | 66.3 % | 70.8 % |
| Digi 1851–1910 tokens | 2.385 G | 69.7 % | 72.7 % |
| Digi up to 1850 types | 3.24 M | 16.0 % | 19.4 % |
| Digi 1851–1910 types | 177.3 M | 3.9 % | 4.9 % |

**Table 2.** Recognition rates for word types and tokens of Digi: OMorfi 0.2 and HisOmorfi

Omorfi 0.2 does not recognize words much better than version 0.1, but HisOmorfi achieves improved recognition of 3 % units with the main part of the data. There is improvement in recognition with HisOmorfi for every type of data, although for word types improvement is small.

[8] https://github.com/jiemakel/omorfi

## 2.2 **Recognition rates decade by decade**

Figure 1. shows recognition rates of words in the data decade by decade without data of 1780–1819 as it consists of Swedish only. Recognition rates are mainly between 65 and 77 per cent. Data of 1770–1779 and 1840–1849 are recognized slightly worse than other data.

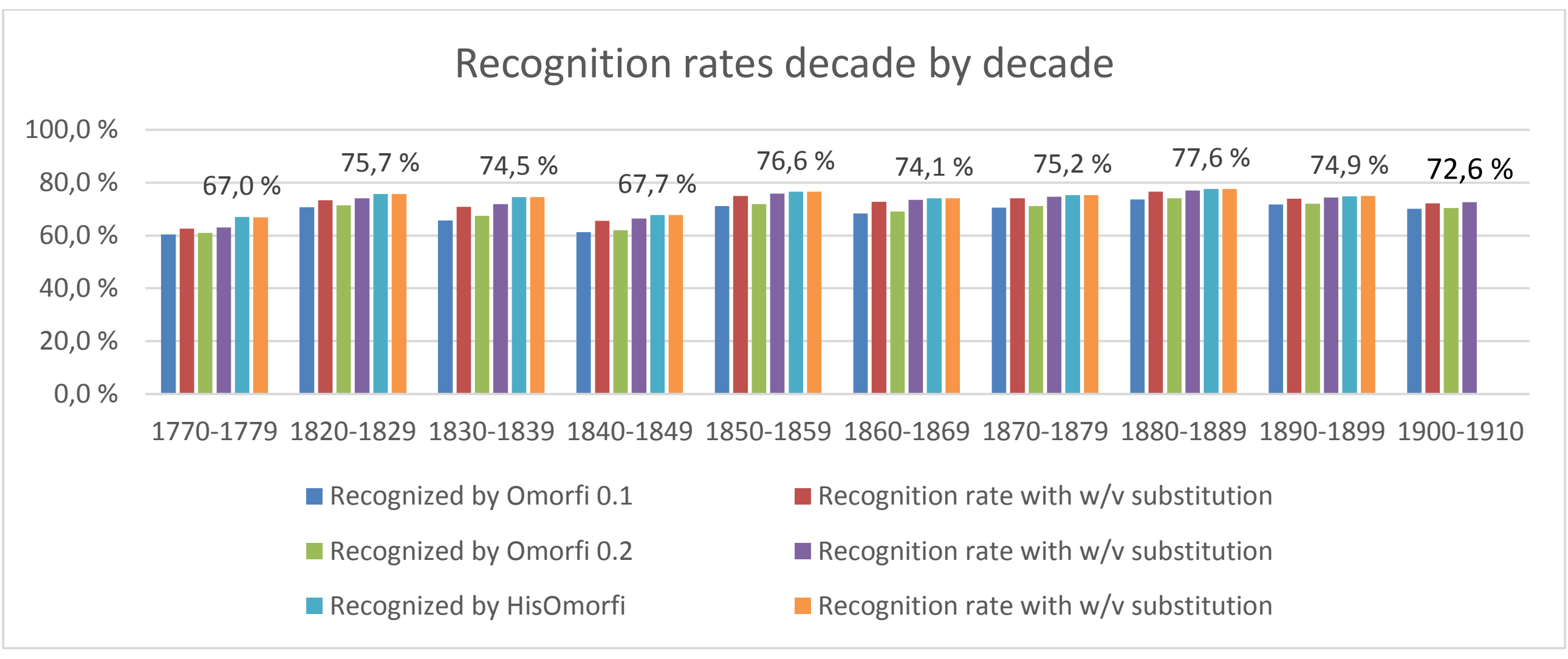

**Figure 1**: Recognition rates of words decade by decade; Data between 1780 and 1819 contains only Swedish and has been omitted. HisOmorfi's recognition rates are given above bars in all other decades but the last, where HisOmorfi's result was not attainable due to character level problems in the data. The last figure is Omorfi 0.2's result with w/v replacement.

Interestingly, there is no big variation in the recognition rates of earlier and late 19th century, although it would be expectable that older data contains more old vocabulary that is not recognized. One reason for quite good recognition of older data may be simpler column structures and larger fonts in older publications, which could have decreased OCR errors. Towards the end of the 19th century number of columns in newspapers increased[9] and also fonts got smaller. Even if Finnish of the late 19th century as such should be easier to recognize for morphological analyzers (cf. also Table 3), it may have more OCR errors due to printing format. We believe these two phenomena have a contrary overall effect on the recognition rate. Also the amount of data may have an effect.

[9] We have gathered layout data related to some of the main newspapers of the 19th century. This data shows that the number of columns went up steeply during 1870-1900. Most of the papers started with two or three columns and ended up to have to five to eight columns at the end of the century.

### 2.3 Analysis of comparable data

At this stage we need also comparable recognition rates for edited lexical data of the same period. For comparison purposes we used all the available material from the Institute for the Languages of Finland. From their web pages[10] we collected two different word corpuses from two different historical periods of Finnish and four different dictionaries from the 19th century. Figures of this data are depicted in Table 3. Sizes of dictionaries refer to unique dictionary entries extracted from the data, not to all of the words in the material. Unless otherwise mentioned, the data consists of word types.

| **Collection** | **Number of words** | **Recognized by Omorfi 0.1** [11] | **Recognized by FINTWOL** | **Type of data** |
|---|---|---|---|---|
| VKS frequency corpus | 285 K | 15 % | 16.6 % | edited[12] mostly 15-18th century material |
| VKS frequency corpus tokens | 3.43 M | 49 % | 50.3 % | edited mostly 15-18th century material |
| VNS frequency corpus [13] | 530 K | 55. 9 % | 58. 1% | edited 19th century material |
| VNS frequency corpus [14] tokens | 4.86 M | 86.1 % | 86.5 % | edited 19th century material |
| Ahlman dictionary 1865 | 91.4 K | 73 % | 71.5 % | edited dictionary material |
| Europaeus dictionary 1853 | 43.2 K | 76 % | 69 % | edited dictionary material |
| Helenius dictionary 1838 | 25.8 K | 49 % | 50 % | edited dictionary material |
| Renvall dictionary 1826 | 25.8 K | 43 % | 45.5 % | edited dictionary material |
| Four dictionaries combined | 132.5 K | 62 % | 61 % | edited dictionary material |

**Table 3**. Historical word data from Institute for the Languages of Finland

We can summarize the recognition rates of the Digi and comparable same period lexical data as a graph depicted in Figure 2.

[10] http://www.kotus.fi/aineistot/tietoa_aineistoista/sahkoiset_aineistot_kootusti

[11] Recognition rates of Omorfi are slightly harmed by the fact that all the data is lower cased and Omorfi recognizes proper nouns from upper case initial letter only

[12] *Edited* here means that the collection was created with manual human labor, by typing the printed version to make it digital.

[13] This material contains the dictionary materials

[14] This material contains the dictionary materials

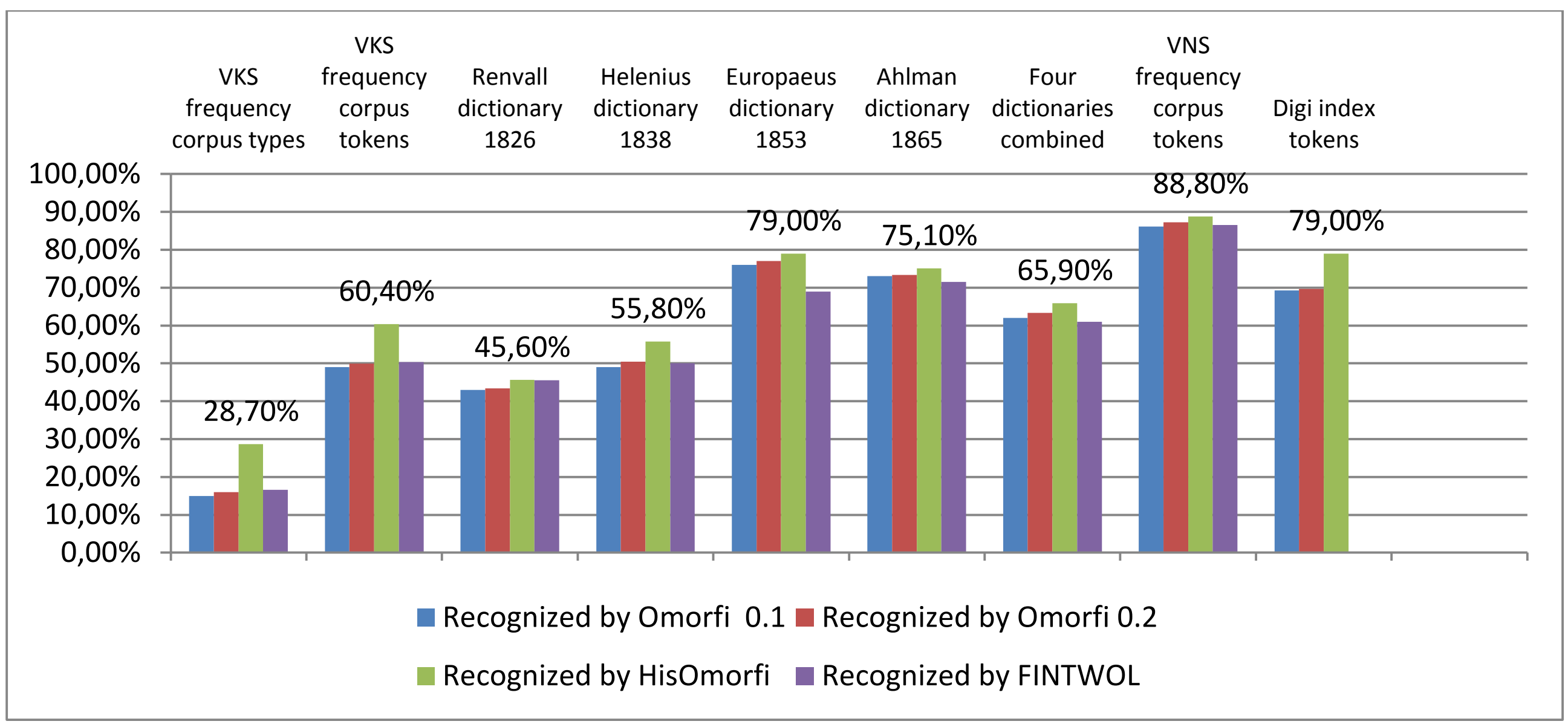

**Figure 2**. Recognition rates of OCRed and edited data. Token level data marked separately, percentages for HisOmorfi's recognition shown in numbers.

As can be seen from Tables 1–3 and Figure 2, type level recognition rates of the Digi data are very low compared to edited comparable material of the same period. The main reason for this is the high number of once occurring words (hapax legomena), most of which are OCR errors, which will be shown in the next chapter (cf. also Ringlstetter et al. 2006). When token level of the Digi data is examined, recognition rates are quite reasonable, 66–79 %. There is a 17–20 % unit difference to the edited comparable data on token level. The exceptionally high recognition rate of the VNS corpus is partly due to the fact that 1000 most frequent types in the corpus consist already 44.6 % of the whole corpus on token level and among these 2.17 M tokens recognition rate is 99.2 %. In the rest of the VNS corpus's 2.69 M of tokens the recognition rate is 76.4 %. Thus a realistic recognition rate for late 19th century data could be around 75–80 % with the tools used.

When we view the results of the morphological analyzers with the edited data, we can also see the clear difference of the historical phase of Finnish between the two larger corpora, VKS and VNS. VKS data is mainly from the period before 19th century (extending to mid-16th century), and thus recognition rates are expectedly much lower. Data in the rest of the corpora are mainly from the 19th century, and their recognition rates are much higher. The two oldest dictionary data, Renvall and Helenius dictionaries, from years 1826 and 1838, get the lowest recognition rates, but even these contain twice or thrice recognized words for the analyzers compared to the VKS data. The two latest dictionaries, Ahlman and Europaeus, get recognition rates of round 70 % and over, four dictionaries combined a rate of about 62 %.

On the basis of the edited data analyses we can approximate, that 56–76 % of the words on type level from the $19^{th}$ century data can be recognized by modern language morphological analyzers. On token level the recognition rate can be up to 80 %, disregarding the effect of most frequent words. If there is older material in the data, recognition will drop, and the drop can be quite large.

### 2.4 Frequency analysis of Digi's data

#### 2.4.1 The most frequent words

Next we proceed to frequency analysis of parts of the data. Table 4. gives results of morphological analyses of 1000–1 M of the most frequent word types of the 1851–1910 part of the Digi word data. At 1 M the frequency of the words is 79, and the frequency range is from 71 257 605 to 79, mean being 2044 and median 194.

Number of unknown words for the analyzers is given on word type level in columns two and three. Column four shows how many tokens of the data each sample covers, and column five shows percentage of the tokens in the whole 2.385 billion words of data.

| **N of word types in the sample** | **Unrecognized word types for Omorfi 0.1** | | **Unrecognized word types for FINTWOL** | | **N of word tokens** | **Coverage of the total data on token level** |
|---|---|---|---|---|---|---|
| **1K** | 134 | 13.4 % | 120 | 12 % | 790 710 542 | 33.1 % |
| **10K** | 1 773 | 17.7 % | 1 767 | 17.7 % | 1 317 532 256 | 55.2 % |
| **100K** | 29 997 | 30 % | 31 457 | 31.5 % | 1 782 767 935 | 74.7 % |
| **500K** | 237 264 | 47.4 % | 245 267 | 49.1 % | 1 983 275 749 | 83.1 % |
| **1 M** | 563 130 | 56.3 % | 577 974 | 57. 8 % | 2 043 976 151 | 85.6 % |

**Table 4.** Morphological analyses of the 1000–1 M of the most frequent word types

For simplicity we analyzed the words on token level with FINTWOL only, and the figures are shown in Table 5.

| N of word types in the sample | N of word tokens | Unrecognized word tokens for FINTWOL | |
|---|---|---|---|
| **1K** | 790 710 542 | 61 170 210 | 7.7 % |
| **10K** | 1 317 532 256 | 152 388 093 | 11.6 % |
| **100K** | 1 782 767 935 | 287 109 856 | 16.1 % |
| **500K** | 1 983 275 749 | 387 237 305 | 19.5 % |
| **1 M** | 2 043 976 151 | 427 214 868 | 20.9 % |

**Table 5.** Number of unrecognized tokens for FINTWOL in 1000–1 M of the most frequent word types

Data in Tables 4 and 5 show that the 1M of the most frequent word types are of quite good quality. On token level only about 21 % of them are unrecognized by FINTWOL, on type level the percentage is about 58 %. The bottom line of Table 5 is that 1.62 G of the tokens of 1 M of the most frequent word types are recognized. Out of the whole data this is 67.6 %. For the rest circa 765 M of tokens the recognition rate is very low, only 30 M of them are recognized.

### 2.4.2 The least frequent words

After analyzing quality of the top of the frequency list, we need to scrutinize the least frequent end of the data. Common to large corpuses are word form types that occur only once in the data, so called *hapax legomena* (Baayen 2001). Number of these in the data is 145 056 481, 81.8 % of the data on type level. Out of these 141 934 402, 97.8 %, are unrecognized by Omorfi 0.1 and 142 221 709, 98 %, are unrecognized by FINTWOL.

In order to confirm occurrence of OCR errors in the least frequent word type classes we analyzed all word types that occur 1–10 times in the data. Table 6. shows the number of these word types, marked as V(m, N), where m is the index for frequency class, N the size of the sample (Baayen 2001, 8). It can be seen that 85–98 % of these word types are unrecognized by the recognizers.

| | N of word types in the sample, V(m, N) | Unrecognized word types for Omorfi 0.1 | | Unrecognized word types for FINTWOL | |
|---|---|---|---|---|---|
| **V(1,N)** | 145 056 481 | 141 934 402 | 97.8 % | 142 221 709 | 98.0 % |
| **V(2,N)** | 13 432 504 | 12 545 050 | 93.4 % | 12 626 341 | 94.0 % |
| **V(3,N)** | 5 223 322 | 4 770 428 | 91.3 % | 4 808 344 | 92.1 % |
| **V(4,N)** | 2 820 741 | 2 536 487 | 89.9 % | 2 558 814 | 90.7 % |
| **V(5,N)** | 1 787 757 | 1 587 586 | 88.8 % | 1 599 055 | 89.4 % |

| **V(6,N)** | 1 240 895 | 1 089 528 | 87.8 % | 1 098 022 | 88.5 % |
|---|---|---|---|---|---|
| **V(7,N)** | 914 598 | 796 136 | 87.0 % | 804 520 | 88.0 % |
| **V(8,N)** | 704 610 | 607 888 | 86.3 % | 614 653 | 87.2 % |
| **V(9,N)** | 560 762 | 480 206 | 85.6 % | 485 741 | 86.6 % |
| **V(10,N)** | 458 734 | 389 775 | 85.0 % | 394 511 | 86.0 % |
| **SUM** | 172 200 404 | 166 737 486 | | 167 211 710 | |

**Table 6.** Number and recognition rates of 1–10 times occurring word types in the data

When we count the number of unknown 2–10 times occurring word types on token level for FINTWOL, we get about 83.4 M words. This added to the number of hapax legomena tokens unrecognized by FINTWOL makes about 225.6 M unrecognized word forms. We believe that these circa 9.4 % of the word form tokens unrecognized by morphological analysis are mostly very hard OCR errors, which are quality wise the worst part of the whole collection (cf. Ringlstetter et al. 2006). The slowly increasing recognition rate among the ten least frequent types suggests that the number of hard OCR errors is somewhere between 225 M and 305 M on token level. Recognition rate at 20 least frequent word types is about 80 %, at 30 about 77 %, at 40 about 75 %, and at 50 about 73 %. Even at 100 the recognition rate is still only about 66 %. Thus the number of unknown word types stays very high at the bottom of the data in a long range.

### 2.5 Orthographical variation and out-of-vocabulary words

Orthography of Finnish was already reasonably stable in the mid-19th century, although there were phenomena that differ from modern language (cf. table 1. in Järvelin et. al 2015). Also dialectical word forms were more common in newspapers of the 19th century. The biggest and most visible difference between modern Finnish and 19th century Finnish is variation of *w/v*, which does not exist in modern language. Thus words that have *w* and are not proper names like *Wien* (*Vienna* in Finnish) are not usually recognized by modern morphological analyzers. To approximate effect of this, we counted the occurrences of *w* in the 1 M of the most frequent words of the data. The data contains 92 749 word types with w, which makes 78 438 010 tokens (3.3 % of all the tokens). Out of the types 91 886 (99.1 %) are unrecognized by FINTWOL. This is 76 450 673 words (97.5 %) on token level. If we replace *w*'s with *v*'s, 54 049 types (58.3 %) are unrecognized by FINTWOL on type level and 24 016 996 (30.6 %) on token level. [15] Thus out of the unrecognized 427 M words of Table 4. 52.4 M

[15] When same analysis is carried out with the 4.86 M tokens of the VNS_Kotus, the original recognition rate of 86.5 improves only to 86.7 %. VNS_Kotus data contains considerably less w's due to the editing policy of the data, cf. http://kaino.kotus.fi/korpus/1800/viite/1800-luvun_korpuksen_koodauksesta.php

(12.2 %) can be recognized with replacement of w. Out of the whole 2.044 G of word tokens of the 1 M of most common types this makes 2.2 %. So effect of the w/v variation among the unrecognized words is significant although the number of the words in all the data is not very high.

To get an approximation of relation between OOV words of the analyzers and OCR errors proper we browsed through the 1 000 most common word types and their 120 unrecognized word types to FINTWOL. Out of these about 85 (70. 8 %) can be considered to be OCR errors, the rest being OOV's. Demarcation out of textual context is not always clear, but we can take the 70 % OCR error figure as a low estimate, and as OCR errors tend to increase with less frequent word types, OCR error percentage could be about 70–90 %.

2.6 **Effect of word length**

It is obvious that length of words plays an important role in OCR of words. We analyzed mean lengths of the 1K–1M range of most common word types and mean lengths of 1-10 least common word types. Results are shown in Figures 3 and 4.

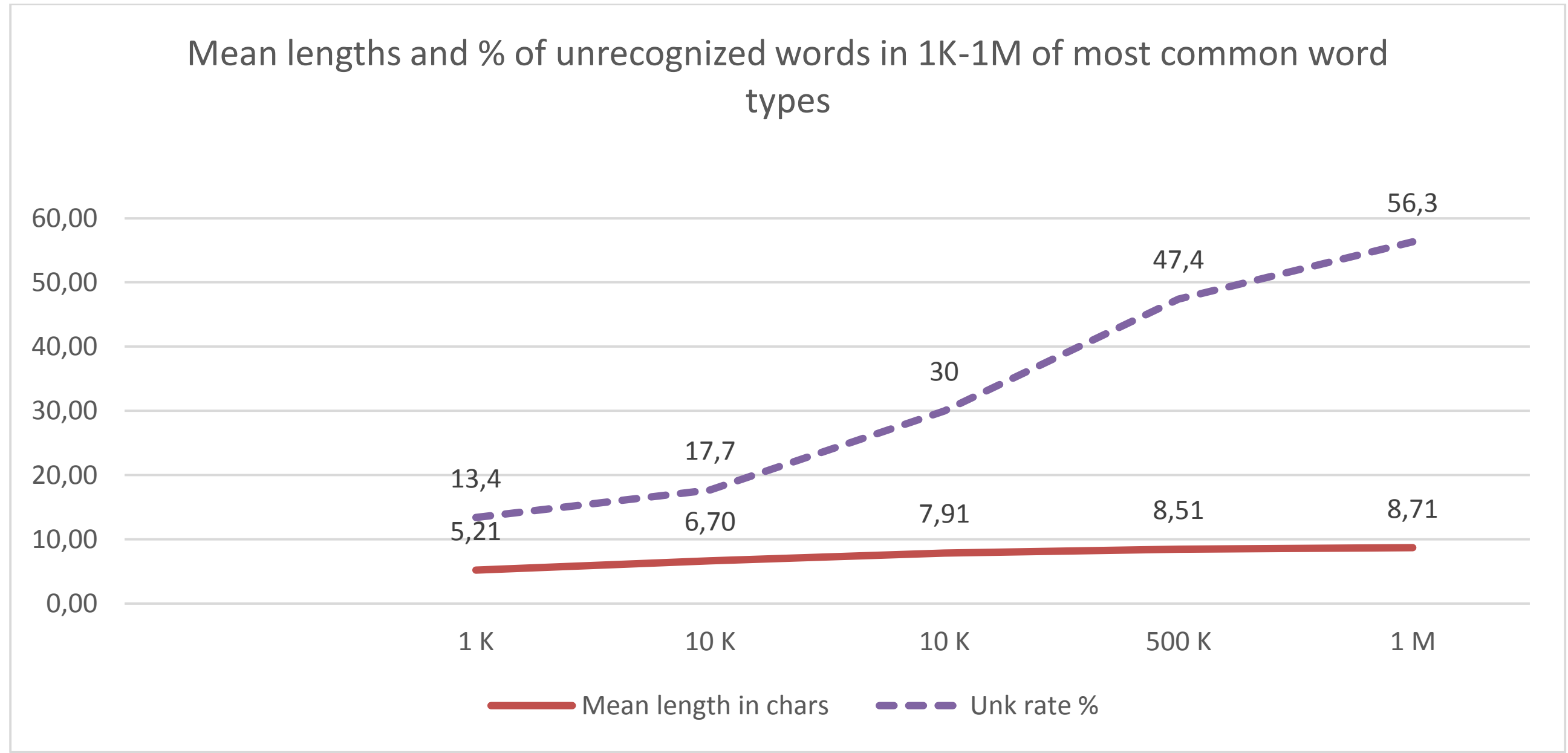

**Figure 3.** Mean lengths and per centages of unrecognized words in 1K–1M of most common word types

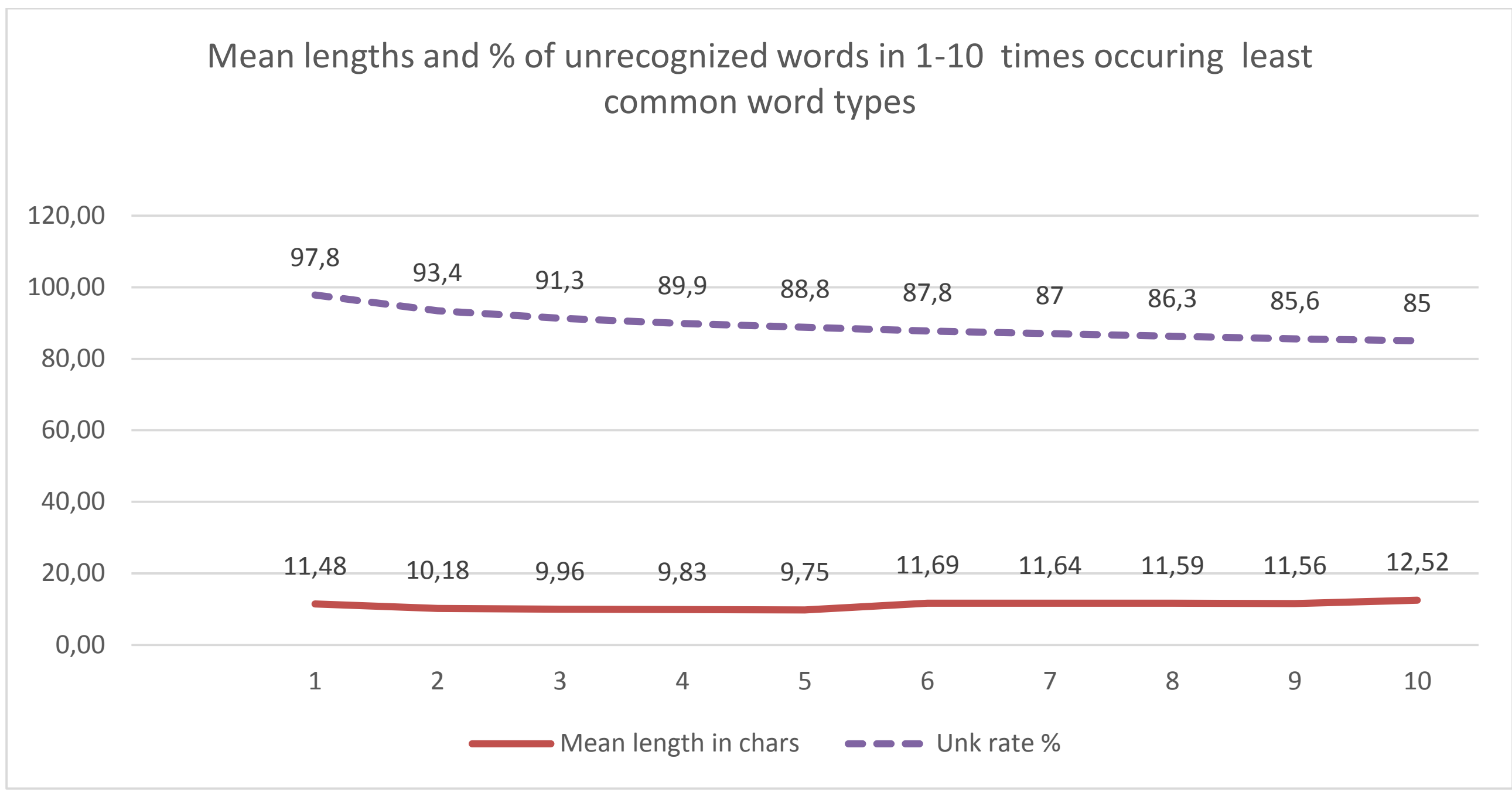

**Figure 4**. Mean lengths and per centages of unrecognized words in 1–10 of least common word types

Figure 3 shows that mean length of words among the 1M most common types grows steadily with clearly decreasing recognition rate. Inverse relation can be seen for least common words: mean length decreases and rate of recognition increases in five least common types, but after that mean length increases again although recognition rate keeps increasing.

### 2.7 **Recognition vs. correctness**

It should, of course, be kept in mind, that recognizability of words is not the same as correctness in the original text. A word may be wrongly OCRed, but still recognizable as a form of some other word. Nonexistent compounds may be recognized, if their composite parts are in the lexicon of the analyzer. As Omorfi has a very large lexicon (424 259 lexemes according to Pirinen, 2015), this may cause lots of false recognitions of compounds. Many words in the Digi's database are split wrongly to parts due to hyphenation in the original text, which may cause both false positive recognition and false negative recognition. Compounds were also written differently in the 19$^{th}$ century Finnish. OOVs, words that are not in the lexicon of the analyzer, bring complexity of their own to results. Some examples of false recognitions and false misrecognitions are shown in the following list:

- *mli* mli Num Roman Nom Sg → probably an OCR error
- *huu* huu Part

*tain* — tai N Gen Sg → wrong division into two parts based on hyphenation, should be *huutain* which is unrecogcnizable, although it is a correct form in the 19th century Finnish
- *Hei* — He Pron Nom Pl → should be *heidan,* unrecognizable *(heidän* would be correct*)*

  *dan* — +?
- *Samoinkuin* — +? → not recognized because written as a compound, correct otherwise
- *ylöskannetaan* — +? → not recognized because written as a compound, correct otherwise

Amount and effect of these kinds of phenomena are hard to estimate, but it is clear that all these phenomena cause uncertainty in the results and make an estimation of error margins in the analysis hard to establish.

## 2.7 **Summing up results of the analysis**

We have now reached a reasonably comprehensive result out of the quality assessment of our data. We have three different parameters that affect the results: number of OOV vocabulary, number of OCR errors proper and effect of *w/v* variation in the data. The effect of the OOV factor in the clean VNS_Kotus data is on token level about 14 % and in the VKS_Kotus about 50 %. Their mean is 32 %, but a fair approximation could be 14–20 % in edited material of the latter part of the 19th century word data. We believe that in the Digi data OCR errors tend to override vastly the OOV words. The variation of w/v has an effect of about 12 % among the unrecognized words.

The initial analysis without considering the w/v variation and effect of OOV's have given us 1.65 G of recognized words and 733 M of unrecognized words (cf. Table 1.). This gives us a figure that we could call *raw recognition rate* of the data, 69 %.

We can now proceed to a more detailed analysis of the 733 M unrecognized words. It is safest to assume, that *w/v* variation and OOV's have most of their effect in the 427 M part of the unrecognized words, because they belong to the most frequent words and constitute 85.6 % of the whole word data. If the number of words recognized when *w*'s are replaced with *v* (52 M) is taken into consideration, the share of recognized words goes up to 71.3 % and share of unrecognized words drops to 28.7 %, absolute number of unrecognized remaining words being circa 375 M. The approximate share of OOV words among the still unrecognized 375 M of words could be somewhere between 50–75 M.

Thus the real number of OCR errors in the data is round 600–625 M, approximately about 25–26 % of the whole. Thus the *approximated recognition rate* of the words in the data could be 74–75 %.

Figure 5 shows how the circa 625 M of unrecognized words are divided. As was shown in Table 5, 225.6 M of the words are from the 1–10 least frequent word types classes and thus they are probably hard OCR error . The same is probably true for the 80 M of unrecognized words found beyond the 1 M of most frequent types and 1–10 least frequent types. The major initial bulk, 427 M, was reduced to about 300 M, and this part is probably easier OCR errors, as the words are among the 1 M of the most frequent types.

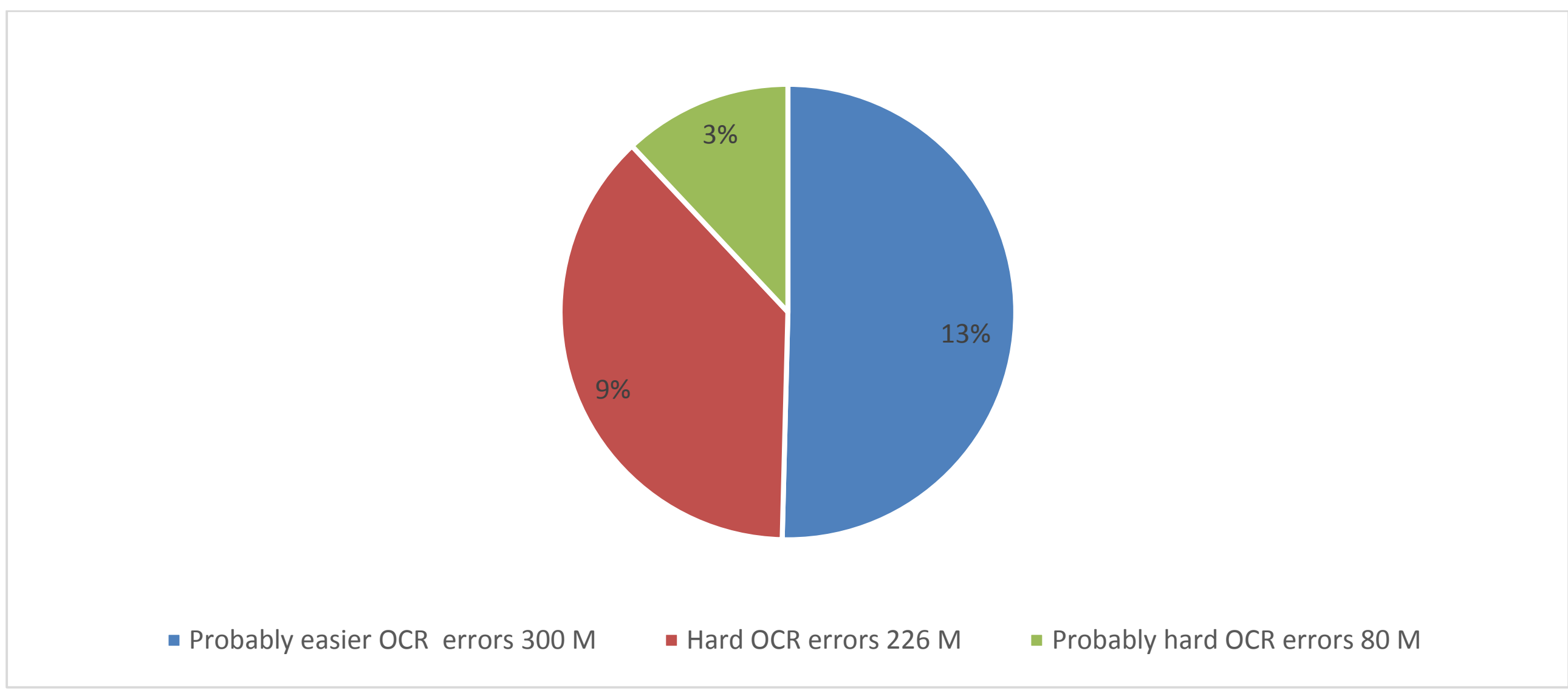

**Figure** 5. Suggested classification of the circa 625 M of unrecognized words

There is no clear way to distinguish easy and hard OCR errors, but suggestive examples are shown in Figure 6 (Kettunen et al., 2014). All the examples derive from the index data of Digi. When the edit distance[16] between the OCR form and correct form increases, the words become unintelligible even for a human native speaker. Such words are e.g. the ones that have edit distance of 5–8. Easier OCR errors could consist of edit distance of 1-4 characters.

[16] https://en.wikipedia.org/wiki/Edit_distance. Informally edit distance is the number of operations needed to transform one string into the other.

| Freq. | OCR form | Correct form | Edit distance | Translation |
|---|---|---|---|---|
| 1 | zzhdysvautki | yhdyspankki | 5 | union bank |
| 1 | zzznuirypäleitä | wiinirypäleitä | 4 | grapes |
| 1 | wiljelystartaltutsessa | wiljelystarkoituksessa | 4 | in a cultivation purpose |
| 1 | urheilutarloinksiin | urheilutarkoituksiin | 3 | for sports purposes |
| 1 | uratkakupoissa | urakkakupoissa (urakkakaupoissa) | 1 | in contract jobs |
| 1 | taitanuiubcsta | taitawuudesta | 4 | of dexterity |
| 1 | taitamattomundestani | taitamattomuudestani | 1 | from my ineptitude |
| 1 | taiötelelutanteren | taistelutanteren | 4 | of the battlefield |
| 1 | taioafliftiutpn | tavallisuuden | 8 | of the usual |
| 1 | taimokkaisuudclllllln | tarmokkaisuudellaan | 6 | with his/her vigor |

**Figure 6.** An illustrative selection of word forms that appear once in the Digi corpus. The correct form is shown with its edit distance to the correct form.

## 3. **Discussion**

Out of all analyses presented, we can make the following conclusions:

- main part, over 99 % of the data in the Digi 1771-1910 is between the years 1851–1910. This implies that the effect of OOV words in the data should not be prohibitively large for analysis with modern language morphological analyzers (Tables 1. and 2.)
- vocabulary of this time period can be recognized reasonably well with modern language morphological analyzers; about 56–76 % of the word types in edited data are recognized, on token level the recognition rate can be as high as 86.5 % (Table 2.)
- in the Digi data, raw token level recognition rate is round 65–69 % (Table 1.) When effect of v/w variation and OOV words is taken into account, the approximated recognition rate is about 74–75 %
- type level recognition rate in Digi data is low (Table 1.), which is due to large set of hapax legomena and other rare word types, that are mainly OCR errors (Table 5.)
- modern Finnish morphological analyzers lack about 30-40 % of the vocabulary of the mainly 19th century edited data on type level; on token level the figure is at minimum about 14 % (Table 2.)

- in up to almost 75 % of the most frequent data (almost 1.8 G of words), raw recognition rate of the word form tokens is about 68.5–70 %. After this the recognition rate drops heavily (Table 3.)
- out of the 625 M of unrecognized words at least about 225 M, 9.4 %, in the data are probably very hard OCR errors, this figure can be up to 313 M (Tables 4. and 5., Figure 6.). About 300 M could be easier OCR errors and perhaps also OOVs (Figure 6.)
- unrecognized words tend to be longer than recognized words (Figures 3. and 4.)

The lexical quality approximation process we have set up is relatively straightforward and does not need complicated tools. It is based on frequency calculations and usage of off-the-shelf modern Finnish morphological analyzers. It can be automatized fully, even if we have done it partly step by step half automatically. It is apparent that we need to be cautious in conclusions, as different data are of different sizes which may cause errors in estimations (Baayen 2001; Kilgariff 2001). However, we believe that our analyses have shed considerable light into quality of the Finnish Digi collection.

At this stage we can also reflect usefulness of the analysis procedure from point of view of improvement of the OCR quality of the Digi collection. The main message that our analysis gives, is that the collection has a relatively good quality part, about 69–75 %, and a very bad quality part, about 9-12 %. The set of about 13 % of the words that are not recognized, is harder to estimate. As a part of them belongs to the most frequent part of the data, they could be at least partly easier OCR errors and OOVs. All in all about a 25–30 % share of the collection needs further processing so that the overall quality of the data would improve.

If correction of the data is performed it should be focused on the 24–25 % unrecognized part of the data. Out of this the ca. 300 M possibly easier part could be improved by post-correction of the material with algorithmic correction software. We have tried post-correction with a sample (Kettunen 2015), but the results were not good enough for realistic post-correction. If post-correction would be focused to only the easier part of the Digi's erroneous data, it could work quite well. General experience from algorithmic post-correction of OCR errors shows, that good quality word material can be corrected relatively well (e.g. Niklas 2010**;** Reynaert 2008). This may also apply to the medium quality word data. But the worst 9–12 % part of the Digi data cannot be corrected with post-processing; only re-OCRing could help with it, as there is so much of it.

Taken that some action had been taken to improve the quality of the Digi data, we have to consider, whether our procedure would be useful in showing quality improvement, if such had been achieved.

We suggest that improvement of the lexical quality could be demonstrated e.g. with following analyses:

- clear improvement in overall recognition rate of the data: at least 3–5 % units in both type and token level analyses[17]
- recognition rate in the top 1 M of the most frequent word types should improve significantly, especially in the 100 K–1 M range, that is now beyond mean recognition rate of edited data
- a very large drop in the number of unrecognized hapax legomena and other rare word types; in practice this would mean tens of millions of word forms to be become recognizable

The major reason for lexical quality assessment of our data is the fact, that OCR errors in the data may have several harmful effects for users of the data. One of the most important effects of poor OCR quality – besides worse readability and comprehensibility - is worse on-line searchability of the documents in the collections (Taghva et al. 1996). Savoy and Naji (2011), for example, analysed how retrieval performance decreases with OCR error corrupted documents quite severely. With mean reciprocal rank as a performance measure, they showed that degradation in retrieval effectiveness is around 17% when dealing with an error rate of 5%. By increasing the error rate to 20%, the average decrease in retrieval is around 46%. Same and larger level of decrease in retrieval effectiveness is shown also in results of the TREC-5's confusion track (Kantor and Voorhees 2000). The effect of errors is not clear cut, however. Tanner et al. (2009) suggest that word accuracy rates less than 80 % are harmful for search, but when the word accuracy is over 80 %, fuzzy search capabilities of search engines should manage the problems caused by word errors. Mittendorf and Schäuble's (2000) probabilistic model for data corruption seems to support this. Information retrieval is robust even with corrupted data, but IR works best with longer documents and long queries. Empirical results of Järvelin et al. (2015) with the Finnish historical newspaper search collection, for example, reveal that even impractically heavy usage of fuzzy matching will help only to a limited degree in search of a low quality OCRed newspaper collection, when short queries and their query expansions are used. Evershed and Fitch (2014), on the other hand, show that if OCR word errors are corrected and word error rate decreased with about 10 % units, recall in document retrieval may have about 9–10 % unit boost with historical OCRed English documents.

[17] We have implemented during the years 2016-2019 a new OCR procedure for our collection using Tesseract's OCR engine 3.04.01. This process is described e.g. in Kettunen and Koistinen (2019). So far we have used the process for re-OCRing one newspaper's (Uusi Suometar) ca. 86 000 pages. The overall improvement measured with recognition rate of HisOmorfi was 15.3 per cent units.

Users of the Digi collection have complained about the poor OCR of the collection relatively little, but some of them have reported curious search results and been annoyed by the OCR quality (Hölttä, 2016; Kettunen, Pääkkönen, Koistinen, 2016). Basing on the empirical search results with the evaluation collection derived from a small subset of the whole Digi material (Järvelin et al., 2016), it is evident that search results in the Digi collection itself are not optimal, and better OCR quality would probably improve them.

Besides retrieval performance effects poor OCR quality has an effect on ranking of the documents (Taghva et al. 1996; Mittendorf and Schäuble 2000). In practice these kinds of drops in retrieval and ranking performance mean that the user will lose relevant documents: either they are not found at all by the search engine or the documents are so low in the ranking list that the user may skip them. Some examples of this in the work of digital humanities scholars are discussed e.g. in Traub et al. (2015).

Weaker searchability of the OCRed collection is only one dimension of poor OCR quality. Other effects of poor OCR quality may show in the more detailed processing of the documents, such as sentence boundary detection, tokenization and part-of-speech-tagging, which are important in higher-level natural language processing tasks (Lopresti 2009). Part of the problems may be local, but part will cumulate in the whole pipeline of NLP processing causing errors. Thus the quality of the OCRed texts is the cornerstone for any kind of further usage of the material, and we need to be able to assess the quality of the data in order to be also able to improve it and show the possible improvements meaningfully.

## 4. Conclusion

In this paper we have suggested how to assess the overall lexical quality of a mainly 19$^{th}$ century OCRed Finnish historical newspaper collection with circa 2.40 billion words. The procedure uses elementary corpus statistics and morphological analyzers of modern Finnish and is straightforward to use. We also propose how to measure quality improvement after correcting the corpus using the suggested procedure.

Advantages of the procedure are the following:

- *coverage*: the procedure gives an approximation of the quality of the whole corpus and in the same time different parts of the whole can be analyzed; thus it is not based on samples only

- *period sensibility*: comparable same period edited lexical corpora are used in assessment and thus the procedure is reasonably sensible to time variation in the data; the method works reasonably well for the so called period of early modern Finnish (ca. 1820-1870) and beginning of modern Finnish (from ca. 1870-), but would be more vulnerable with earlier material, as lexical coverage of the morphological recognizers would be lower
- *simplicity*: available modern language technology tools and basic corpus statistic methods are used, and no high-level tools need to be developed;

The main vulnerability of the proposed procedure at present is possible sampling error and its effects with corpora used. This, however, can be taken into account with adding advanced statistics to the procedure. They may sharpen the procedure, but at present we are satisfied with the current approach and believe that the measures the procedure produces are useful in quality assessment and quality control after improvements in the word data.

## Acknowledments

This work was supported by European Regional Development Fund (ERDF), Leverage from the EU, 2014–2020.

## Word list sources